\documentclass[sigconf]{acmart}

\usepackage{booktabs} 
\usepackage[ruled]{algorithm2e} 
\usepackage{flushend} 
\usepackage{multirow}
\usepackage{amsmath,amssymb} 
\usepackage{algorithmicx}

\setcopyright{acmlicensed}
\usepackage{subcaption}
\usepackage{xspace}

\makeatletter
\DeclareRobustCommand\onedot{\futurelet\@let@token\@onedot}
\def\@onedot{\ifx\@let@token.\else.\null\fi\xspace}

\def\eg{\emph{e.g}\onedot}

\def\etc{\emph{etc}\onedot} 
 
\def\etal{\emph{et al}\onedot}
\makeatother

\acmDOI{10.1145/3240508.3240660}

\acmISBN{978-1-4503-5665-7/18/10}

\acmConference[MM '18]{2018 ACM Multimedia Conference}{October 22--26, 2018}{Seoul, Republic of Korea} 
\acmYear{2018}
\copyrightyear{2018}

\acmPrice{15.00}
\begin{document}
\title{Adaptive Temporal Encoding Network for Video Instance-level Human Parsing}

\author{Qixian Zhou}
\affiliation{%
  \institution{Sun Yat-sen University}
}
\email{qixianzhou.mail@gmail.com}

\author{Xiaodan Liang}
\authornote{Corresponding author.}
\affiliation{%
  \institution{Carnegie Mellon University}
}
\email{xdliang328@gmail.com}

\author{Ke Gong}
\affiliation{%
  \institution{Sun Yat-sen University}
}
\email{gongk3@mail2.sysu.edu.cn}

\author{Liang Lin}
\affiliation{%
  \institution{Sun Yat-sen University \& SenseTime Research}
}
\email{linliang@ieee.org}


\begin{abstract}
Beyond the existing single-person and multiple-person human parsing tasks in static images, this paper makes the first attempt to investigate a more realistic video instance-level human parsing that simultaneously segments out each person instance and parses each instance into more fine-grained parts (\eg, head, leg, dress). We introduce a novel Adaptive Temporal Encoding Network (ATEN) that alternatively performs temporal encoding among key frames and flow-guided feature propagation from other consecutive frames between two key frames. Specifically, ATEN first incorporates a Parsing-RCNN to produce the instance-level parsing result for each key frame, which integrates both the global human parsing and instance-level human segmentation into a unified model. To balance between accuracy and efficiency, the flow-guided feature propagation is used to directly parse consecutive frames according to their identified temporal consistency with key frames. On the other hand, ATEN leverages the convolution gated recurrent units (convGRU) to exploit temporal changes over a series of key frames, which are further used to facilitate the frame-level instance-level parsing. By alternatively performing direct feature propagation between consistent frames and temporal encoding network among key frames, our ATEN achieves a good balance between frame-level accuracy and time efficiency, which is a common crucial problem in video object segmentation research. To demonstrate the superiority of our ATEN, extensive experiments are conducted on the most popular video segmentation benchmark (DAVIS) and a newly collected Video Instance-level Parsing (VIP) dataset, which is the first video instance-level human parsing dataset comprised of 404 sequences and over 20k frames with instance-level and pixel-wise annotations.

\end{abstract}


%
%

\maketitle

\section{Introduction}
Human parsing is the task of recognizing multiple semantic parts (\eg, head, legs), which is a fundamental and significant problem that has received increasing attention, because of its significant potential in more high-level applications (\eg, human behavior analysis~\cite{gan2016concepts,liang2015proposal,liang2016reversible,liang2017deep,cao2018visual,liang2017dual}, video surveillance~\cite{wang2014deformable,zhang2016faster,li2017perceptual,li2018scale}).

Due to the successful development of fully convolutional networks (FCNs)~\cite{long2014fully}, great progress has been made in human parsing, or semantic part segmentation task~\cite{chen2016deeplab,ATR,crfasrnn,Co-CNN,Gong_2017_CVPR,liang2016semantic,liang2017interpretable,liang2018dynamic}. However, previous approaches for single-person or multiple-person human parsing only focus on the static image domain. Towards the research closer to real-world scenarios, fast and accurate video instance-level human parsing is more desirable and crucial for high-level applications such as action recognition and object tracking as well as group behavior prediction.

\begin{figure}[t]
\centering
  \includegraphics[width=0.9\linewidth]{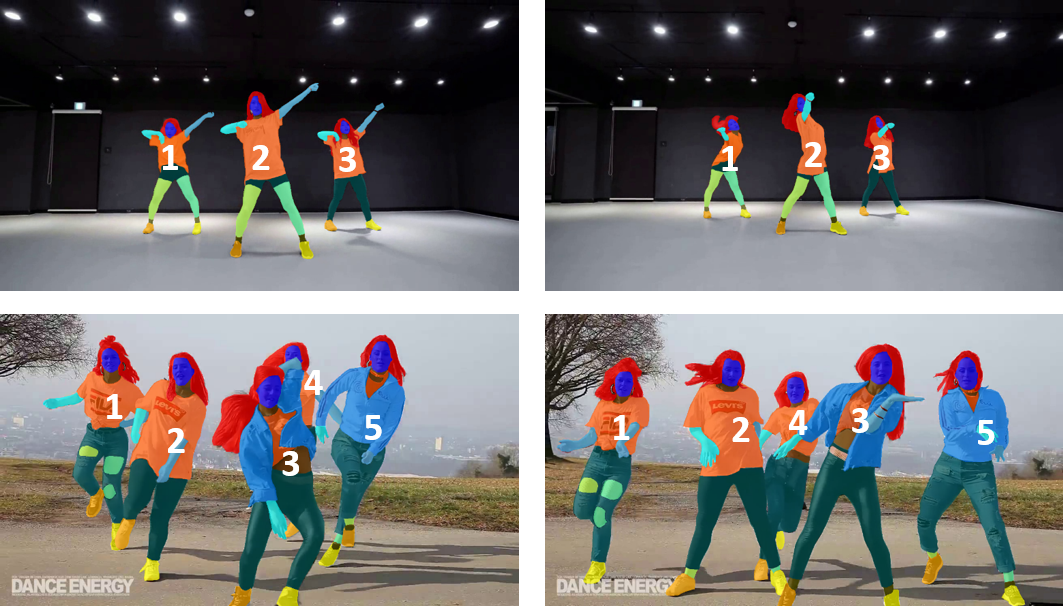}
\vspace{-4mm}
\caption{Sample sequences from our VIP dataset, with ground truth part segmentation masks overlaid. }
\vspace{-4mm}
\label{fig:vip_dataset}
\end{figure}

In this work, we make the first attempt to investigate the more challenging video instance-level human parsing task, which needs to not only segment various body parts or clothes but also associate each part with one instance for every frame in the video, as shown in Fig.~\ref{fig:vip_dataset}. Besides the difficulties shared with single-person parsing (\eg, various appearance/viewpoints, self-occlusions) and instance-level parsing (\eg, uncertain quantities of instances), video human parsing faces more challenges that are inevitable in video object detection and segmentation problems. For example, recognition accuracy suffers from deteriorated appearances in videos that are seldom observed in still images, such as motion blur and video defocus. On the other hand, the balance between frame-level accuracy and time efficiency is also very difficult and important factor to the deployment of diverse devices (such as mobile).

Recently, there have been lots of attempts to build CNNs for video domain vision recognition (\eg, video object detection, video semantic segmentation)~\cite{Zhu2017Deep,Zhu2017Flow,Jin2017Video,Shelhamer2016Clockwork}. Among them, feature-level approaches~\cite{Zhu2017Deep,Shelhamer2016Clockwork,Zhu2017Flow} have drawn a lot of attention. These approaches divide the image domain methods into two steps: 1) obtaining the image feature by deep convolution network; 2) the result is generated subsequently by task-specific sub-network. Any image domain methods can be transferred to video domain task by enhancing the feature of images via embedding temporal information. Moreover, to decrease the computation cost induced by processing numerous video frames, feature-level approaches reuse the sparse key frame by optical flow warping for the temporal redundant information between nearby frames. However, these approaches obtain high speed at the expense of the model accuracy. On the other hand, some approaches~\cite{Zhu2017Flow} improve accuracy by aggregating nearby frame feature to strengthen low quality feature which caused by object motion blur. This operation enables a high performance in video tasks but sacrifices speed because multi-frame aggregation is time-expensive for redundant feature propagation.

In this work, we introduce a new framework, called Adaptive Temporal Encoding Network(ATEN), to improve video task accuracy and keep high inference speed simultaneously. 
First, we split videos into segments with equal length. Each segment only has one key frame which would pass through deep convolution network to extract features. Inspired by flow-guided feature propagation~\cite{Zhu2017Deep}, extracted features will be propagated from key frames to nearby non-key frames via flow field. Because of the temporal redundant between the consecutive frames, it is unnecessary to do frame feature computation densely. Applying flow-guided feature propagation, our approach is more time-efficient as the process of the optical flow estimation and the feature propagation are much faster than passing each frame into backbone network to extract feature. To fully utilize key frame feature, we assign the middle frame of one segment as the key frame, which produces the shortest average distance between key frame and non-key frames. 

To further improve frame-level accuracy, we leverage the convolution gated recurrent units (convGRU) to exploit temporal changes over a series of key frames and enhance low-quality key frame feature caused by motion blur. In our framework, we employ DeeplabV3+ model~\citep{Chen2018Encoder} as our backbone network for feature extraction and a pre-trained FlowNet~\cite{Dosovitskiy2015FlowNet} to estimate optical flow. Our ATEN learns to enhance key frame feature by incorporating a flow-guided feature warping followed by temporal gated recurrent units encoding. The output feature maps are fed into our novel Parsing R-CNN for instance-level human parsing task. Parsing R-CNN is a unified framework extending Mask R-CNN\citep{He_2017_ICCV} with a fine-grained global human parsing branch. It segments instance-level human segmentation via Mask R-CNN branch and predicts semantic fine-grained part segmentation via global human parsing branch simultaneously. Then instance-level part maps are generated by aggregation human segmentation maps and part segmentation maps, which become the results of instance-level human parsing task. All the three modules of ATEN including flow-guided feature propagation, temporal key feature encoding as well as Parsing R-CNN can be trained end-to-end as a unified network.

Furthermore, to facilitate more advanced human analysis, we first establish a standard representative benchmark aiming to cover a wide pallet of challenges for the video instance-level human parsing task. We propose a new large-scale dataset named as ``Video Instance-level Parsing (VIP)'', which contains 404 multiple-person videos and over 20k frames. The videos are high-resolution Full HD sequences where the images are pixel-wisely annotated with 19 semantic parts in instance-level.

Our contributions are summarized in the following aspects. 1) We make the first attempt to thoroughly investigate the challenging video instance-level human parsing task, which pushes the research boundary of human parsing from static images to temporal sequences. 2) We present a novel Adaptive Temporal Encoding Network (ATEN) that alternatively performs temporal encoding and flow-guided parsing propagation and achieves high frame-level accuracy as-well-as time efficiency. 3) We establish the first large-scale video instance-level human parsing dataset with fine-grained annotations. 4) The proposed ATEN achieves state-of-the-art performance for both video instance-level human parsing and video object segmentation on our new VIP dataset and the public DAVIS dataset~\cite{perazzi2016benchmark}.

\begin{figure*}[t]
\centering
  \includegraphics[width=0.8\linewidth]{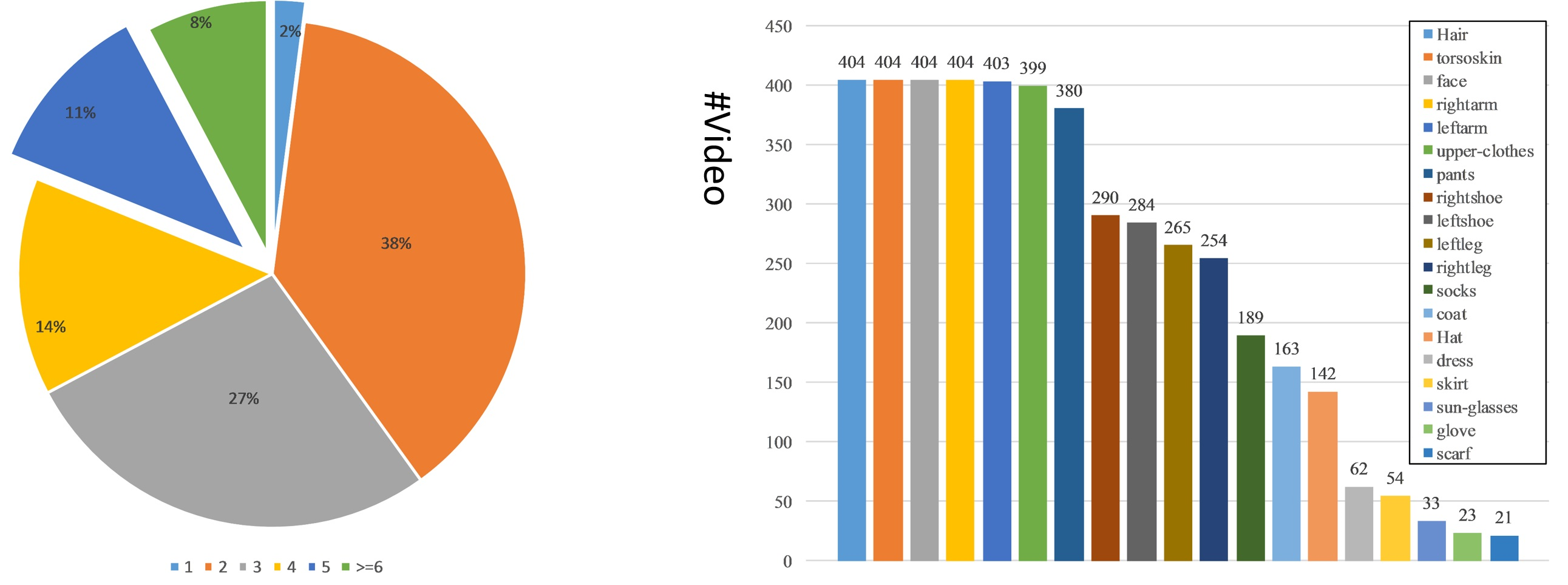}
\vspace{-4mm}
\caption{Left: Statistics on the number of persons in one frame. Right: The data distribution on 19 semantic part labels in the VIP dataset.}
\vspace{-4mm}
\label{fig:vip_statistic}
\end{figure*}

\section{Related Work}
\textbf{Human Parsing}
Driven by the advance of fully convolutional networks (FCNs)~\cite{long2014fully}, more and more research efforts have been devoted to human parsing~\cite{Co-CNN,yamaguchi2012parsing,Yamaguchiparsing13,liang2018look,liang2016semantic,SimoSerraACCV2014,M-CNN,xia2015zoom,chen2015attention,Gong_2017_CVPR}. For example, Liang \etal.~\cite{Co-CNN} proposed a novel Co-CNN architecture that integrates multiple levels of image contexts into a unified network. Gong \etal~\cite{Gong_2017_CVPR} presented a structure-sensitive learning to enforce the produced parsing results semantically consistent with the human joint structures. Liu \etal~\cite{Liu2017Surveillance} introduced a Single frame Video Parsing (SVP) method under video surveillance scene, which requires only one labeled frame per video in training stage. However, all these prior works only focus on the relatively simple single-person human parsing in the still images without considering the more common multiple instance cases in the real world. One of the most important reason is that previous datasets~\cite{yamaguchi2012parsing,Co-CNN,Gong_2017_CVPR,chen2014detect} only include very few person instances and categories in the still images, and require prior works only evaluate pure part segmentation performance while disregarding their instance belongings. On the contrary, containing 404 multiple-person videos and over 20k frames, the proposed VIP dataset is the first fine-grained dataset for video instance-level human parsing, which can further facilitate the human analysis research.

\textbf{Video Object Segmentation}
There is lots of significant progress~\cite{faktor2014consensusvoting,Papazoglou2014Fast,Taylor2015Causal,TokmakovICCV2017,Bideau2016It,TokmakovCVPR2017,han2018reinforcement,yuan2017temporal} in video object segmentation due to the success of deep convolution network. Most of the  effective methods~\cite{faktor2014consensusvoting,Papazoglou2014Fast,Taylor2015Causal,TokmakovICCV2017} formulate this problem as a binary classification problem. These methods generate the final object segmentation by integrating appearance module and temporal object motion capture module (\eg, pairwise constraints~\cite{Papazoglou2014Fast}, occlusion relations~\cite{Taylor2015Causal}, recurrent neural network~\cite{TokmakovICCV2017}). A very recent work~\cite{TokmakovICCV2017} learns the appearance feature via a deep fully convolutional network and object motion feature via specific motion network. A convolutional gated recurrent unit is used to model temporal coherence among nearby frames with the concatenate feature of appearance feature and motion feature. However, this method fails to achieve time efficient as it not only computes deep convolution feature for each frame but also densely updates appearance and motion feature via convGRU with several nearby frames. Another attempt~\cite{Bideau2016It,TokmakovCVPR2017} to solve video object segmentation regards this as the task of segmenting objects in motion, irrespective of camera motion. MP-Net~\cite{TokmakovCVPR2017} learns to segment moving object in a video with only optical flow input. Nevertheless, these methods still lack strong robustness with difficulty to segment some salience object without moving. Superior to the previous methods, our ATEN achieves a good balance between frame-level accuracy and time efficiency by alternatively performing direct feature propagation between consistent frames and temporal encoding network among key frames. 

There exist several datasets for video object segmentation, but none of them has been specifically designed for video instance-level human parsing. The MoSeg dataset~\cite{brox2010object} is a popular dataset for motion segmentation but most of the videos have low spatial resolution, segmentation is only provided on a sparse subset of the frames. The Berkeley Video Segmentation Dataset (BVSD)~\cite{sundberg2011occlusion} comprises higher resolution but a total 100 sequences. DAVIS~\cite{perazzi2016benchmark} is a newly developed dataset for video object segmentation, which contains 50 high quality and full HD video sequences with 3,455 densely annotated pixel-level and per-frame ground-truth. Containing 404 multiple-person videos and over 20k frames, our VIP dataset is the largest fine-grained video parsing dataset focusing on video instance-level human parsing.

\begin{figure*}
  \centering
  \includegraphics[width=.8\textwidth]{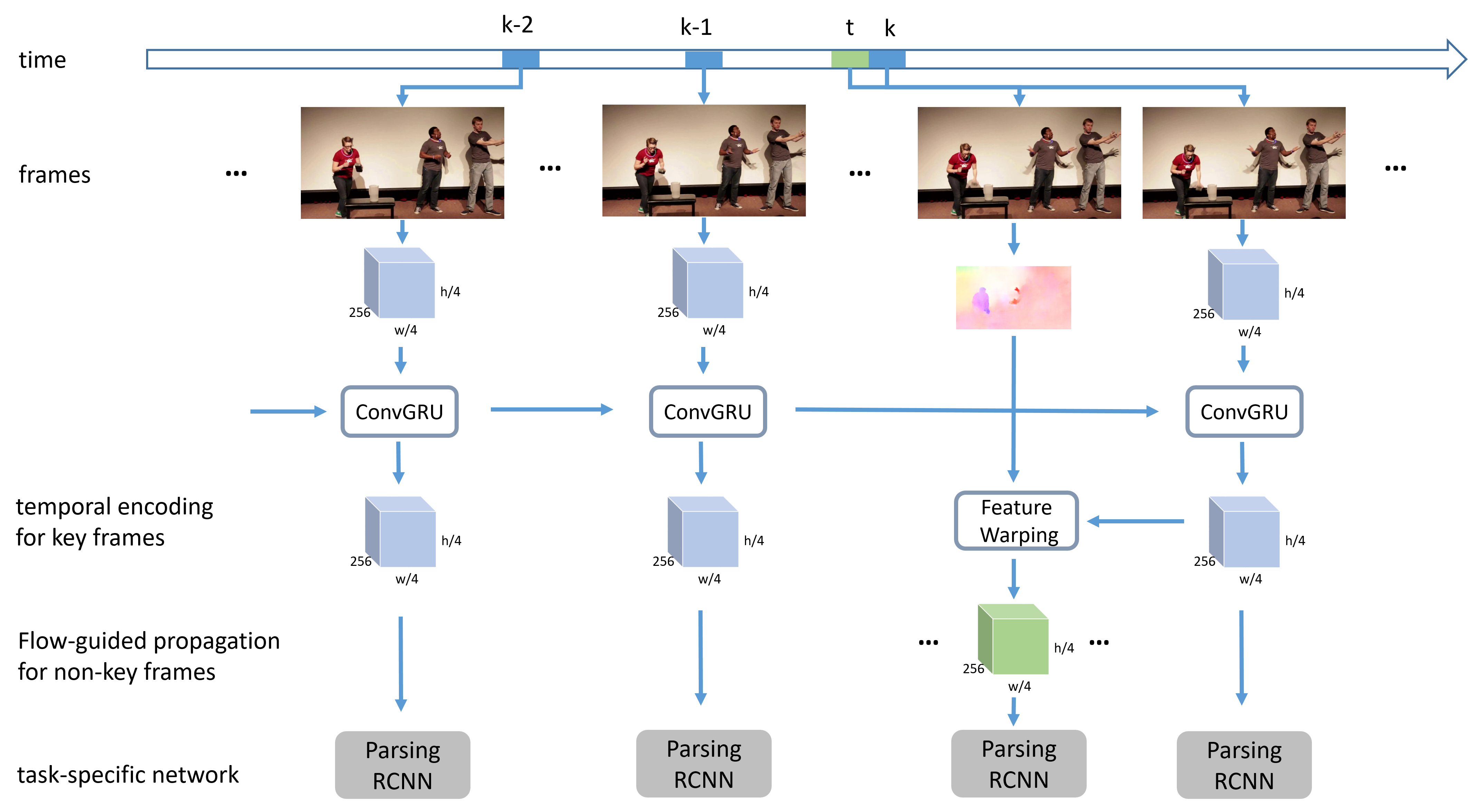}
\vspace{-4mm}
\caption{An overview of our ATEN approach, which performs adaptive temporal encoding over key frames and flow-guided feature propagation for consecutive frames among key frames. Each key frame (blue) is fed into a temporal encoding module that memorized the temporal information of its former key frames. To alleviate the computation cost, the features of consecutive frames (green) among two key frames can be produced by flow-guided propagation module from its nearest key frame. After that, all feature maps of all frames are fed to Parsing-RCNN to generate the instance-level human parsing results.}
  
\vspace{-4mm}
\label{fig:1}
\end{figure*}

\section{Video Instance-level Parsing Dataset}
In this section, we describe our new Video Instance-level Parsing (VIP) dataset\footnote{The dataset is available at http://sysu-hcp.net/lip} in detail. Example frames of some of the sequences are shown in Fig.~\ref{fig:vip_dataset}. To the best of our knowledge, our VIP is the first large-scale dataset that focuses on comprehensive human understanding to benchmark the new challenging video instance-level fine-grained human parsing task. Containing the videos collected from the real-world scenarios where people appear with various poses, viewpoints and heavy occlusions, the VIP dataset involves the difficulties of the semantic part segmentation task. Furthermore, it also includes all major challenges typically found in longer video sequences, like motion blur, camera shake, out-of-view, scale variation, \etc.

\subsection{Data Amount and Quality} 
Our data collection and annotation methodology were carefully designed to capture the high variability of real-world human activities scenes. The sequences are collected from Youtube with several specified keywords (e.g., dancing, flash mob, etc.) to gain a great diversity of multi-person videos. All images are annotated meticulously by the professional workers. We maintain data quality by manually inspecting and conduct a second round check for annotated data. We remove the unusable images that are of low resolution and image quality. The length of a video in the dataset ranges from 10 seconds to 120 seconds. For every 25 consecutive frames in each video, one frame is annotated densely with pixel-wise semantic part categories and instance-level identification. 

\subsection{Dataset Statistics}
To analyse every detailed region of a person including different body parts as well as different clothes styles, following the largest still image human parsing LIP dataset~\cite{Gong_2017_CVPR}, we defined 19 usual clothes classes and body parts for annotation, which are Hat, Hair, Sunglasses, Upper-clothes, Dress, Coat, Socks, Pants, Gloves, Scarf, Skirt, Torso-skin, Face, Right/Left arm, Right/Left leg, and Right/Left shoe. The numbers of videos for each semantic part label are presented in Fig.~\ref{fig:vip_statistic} (Right). Additionally, superior to the previous attempts~\cite{Co-CNN,Gong_2017_CVPR,chen2014detect} with average one or two person instances in an image, the annotated frames of our VIP dataset contain more instances with an average of 2.93. The distribution of the number of persons per frame is illustrated in Fig.~\ref{fig:vip_statistic} (Left).

\section{Adaptive Temporal Encoding Network}
\label{sess:methodoverview}
Given a video frame sequence $I_{j},j=1,2,3...,N$, the video instance-level human parsing is to output each person instance and parse each instance into more fine-grained parts(\eg, head, leg, dress) of all frames. A baseline approach to solving this problem is applying an image instance-level human parsing method on each frame individually, which is simple but has poor performance (efficiency and accuracy) because of the lack of temporal information. Firstly, as a baseline, we propose a novel Parsing-RCNN to produce the instance-level parsing result for each key frame, which integrates both the global human parsing and instance-level human segmentation into a unified model. In Parsing-RCNN, a deep fully convolutional network(FCN) is applied on the input image $I$ to generate feature maps $F = \mathcal{N}_{feat}(I)$. Subsequently, a well designed instance-level human parsing sub-network $\mathcal{N}_{parse}$ is applied on the extracted features to produce global human parsing as well as instance-level human segmentation, and generate the final instance-level human parsing results $R = \mathcal{N}_{parse}(F)$ by taking the union of all parts assigned to a particular instance.

As shown in Fig \ref{fig:1}, our ATEN approach based on Parsing-RCNN aims at balancing both efficiency and accuracy by applying flow-guided feature propagation and adaptive temporal encoding. We split each video sequence into several segments of equal length $l$, $Seg = [I_{jl}, I_{jl+1}, ..., I_{(j+1)*l-1}]$. Only one frame in each segment is selected to be a key frame (using the median frame as default). Given a key frame $I_{k}$, the encoded feature is denoted as 
\begin{equation}
\overline{F}_{k} = \varepsilon(F_{k-2}, F_{k-1}, F_{k})
\end{equation}
Subsequently, The feature of a non-key frame $I_{t}$ are propagated from nearest key frame $I_{k}$, which is denoted as 
\begin{equation}
\overline{F}_{t} = \mathcal{W}(\overline{F}_{k}, M_{t{\to}k}, S_{t{\to}k})
\end{equation}
where $M$ and $S$ is the flow field and scale field respectively. Finally, the instance-level human parsing sub-network $\mathcal{N}_{parse}$ is applied on both encoded key frame feature maps and warped non-key frame feature maps to compute eventual result $R = \mathcal{N}_{parse}(\overline{F})$.

\begin{figure}
  \includegraphics[width=.45\textwidth]{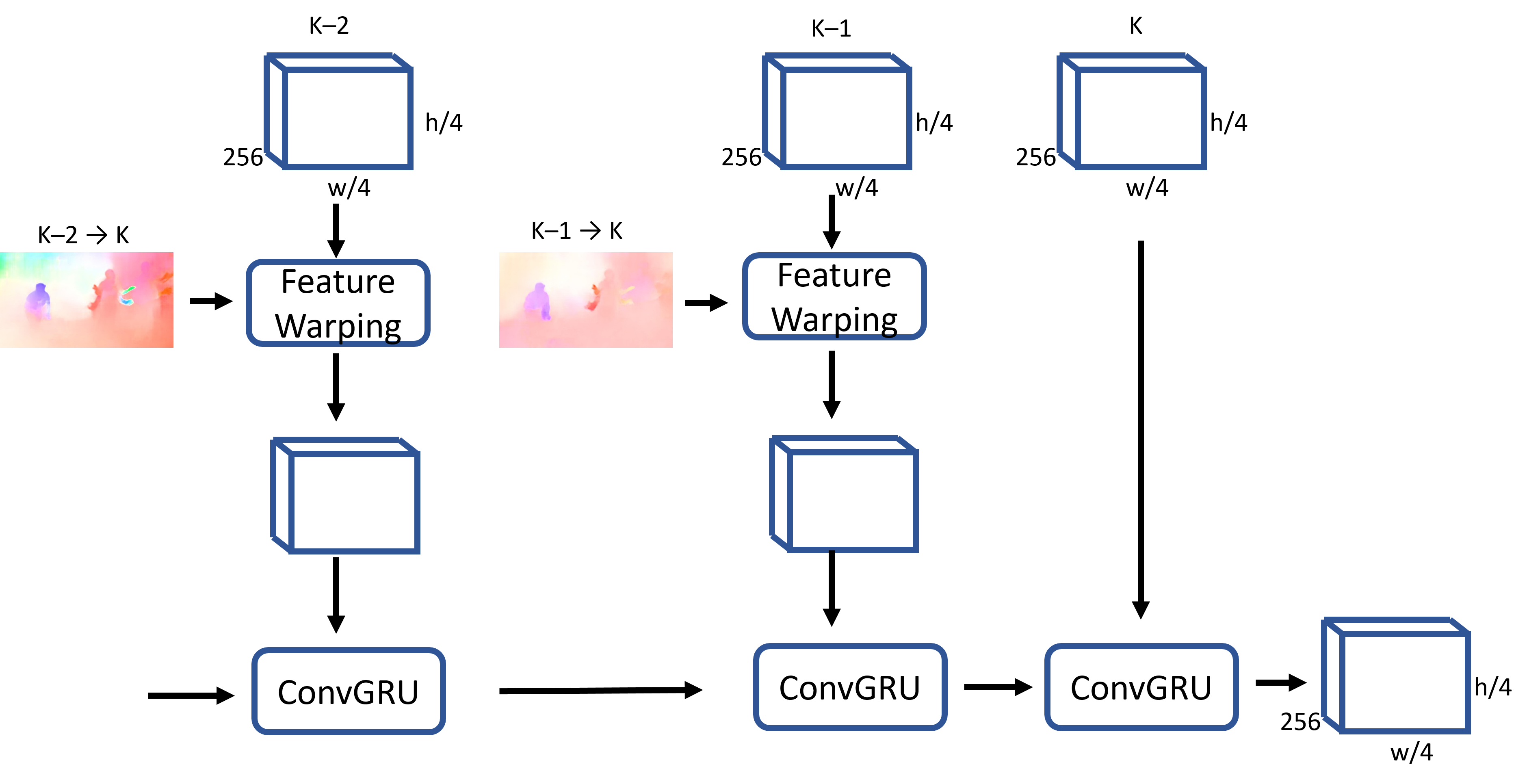}
\vspace{-4mm}
  \caption{Our adaptive temporal encoding module. For each key frame $K$, we first obtained warped feature maps from two previous key frames (i.e. $K-1$ and $K-2$) via flow-guided propagation module. Then the warped features and current appearance features are consecutively fed to convGRU for temporal encoding. All feature maps in this module have the same shape (stride of 4, 256 dimensions).}
\vspace{-4mm}
  \label{fig:2}
\end{figure}

\subsection{Adaptive Temporal Encoding}
\label{sess:ate}
As shown in Fig \ref{fig:2}, given a encoding range $p$ which specifies the range of the former key frames for encoding ($p$ = 2 by default), we first apply the embedded FlowNet $\mathcal{F}$\cite{Dosovitskiy2015FlowNet} to individually estimate $p$ flow fileds and scale fileds, which are used for warping (as illustrated in Session \ref{sess:flow}) $p$ former key frames to current key frame. 
\begin{equation}
F_{k-j{\to}k} = \mathcal{W}(F_{k-j}, M_{k{\to}k-j}, S_{k{\to}k-j}), j{\in}[1, p]
\end{equation}
After feature warping, each warped feature is consecutively fed to convGRU for temporal coherence feature encoding. We use the last state of GRU as the encoded feature.
\begin{equation}
\overline{F}_{k} = convGRU(F_{k-p{\to}k}, ..., F_{k-1{\to}k}, F_{k})
\end{equation}

ConvGRU is an extension of traditional GRU\cite{Cho2014Learning} which has convolutional structures instead of fully connected structures. Equation(\ref{con:gru_equation}) illustrates the operations inside a GRU unit. The new state $h_t$ is a weighted combination of the previous state $h_{t-1}$ and the candidate memory $h'_t$. The update gate $z_t$ determines how much of this memory is incorporated into the new state. The reset gate $r_t$ controls the influence of the previous state $h_{t-1}$ on the candidate memory $h'_t$.
\begin{equation}
\begin{aligned}
&z_t = \sigma(x_t * w_{xz} + h_{t-1} * w_{hz} + b_z),\\
&r_t = \sigma(x_t * w_{xr} + h_{t-1} * w_{hr} + b_r),\\
&h'_t = tanh(x_t * w_{xh'} + r_t {\odot} h_{t-1} * w_{hh'} + b_{h'}),\\
&h_t = (1 - z_t) {\odot} h_{t-1} + z_t {\odot} h'_t,
\end{aligned}\label{con:gru_equation}
\end{equation}

Unlike traditional GRU, $*$ here represents a convolutional operation. ${\odot}$ denotes element-wise multiplication. $\sigma$ is the sigmoid function. $w$ are learned transformations and $b$ are bias terms.

\subsection{Flow-guided Feature Propagation}
\label{sess:flow}
Motivated by \cite{Zhu2017Deep, Zhu2017Flow}, given a reference frame $I_{j}$ and target $I_{i}$ frame, a optical flow field is calculated by embedded FlowNet $\mathcal{F}$\cite{Dosovitskiy2015FlowNet, Ilg2017FlowNet} to obtain pixel-wise motion path. Extending the FlowNet with scale field which is of the same spatial and channel dimensions as the feature maps helps to improve flow warping accuracy. The feature propagation function is defined as:
\begin{equation}
F_{j{\to}i} = \mathcal{W}(F_j, {\mathcal{F}}(I_i, I_j)){\odot}S
\end{equation}
where $F_j$ denotes the deep feature of reference frame $I_{j}$. $\mathcal{W}$ denotes bilinear sampler function. ${\odot}$ denotes element-wise multiply. ${\mathcal{F}}$ represents flow estimation function and $S$ is scale field that refines the warped feature. FlowNet-S\cite{Dosovitskiy2015FlowNet} are adopt as flow estimation function and pre-trained on FlyingChairs dataset. A scale map with the same dimensions as target features is predicted in parallel with flow field by FlowNet via an additional $1{\times}1$ convolution layer attached to the top feature of the flow network. The weights of extra $1{\times}1$ convolution layer are initialized with zeros. The biases are initialized with ones and frozen during train phase. The whole process is fully differentiable, which has been well described in \citep{Zhu2017Deep, Zhu2017Flow}. 

\begin{figure}
  \includegraphics[width=.45\textwidth]{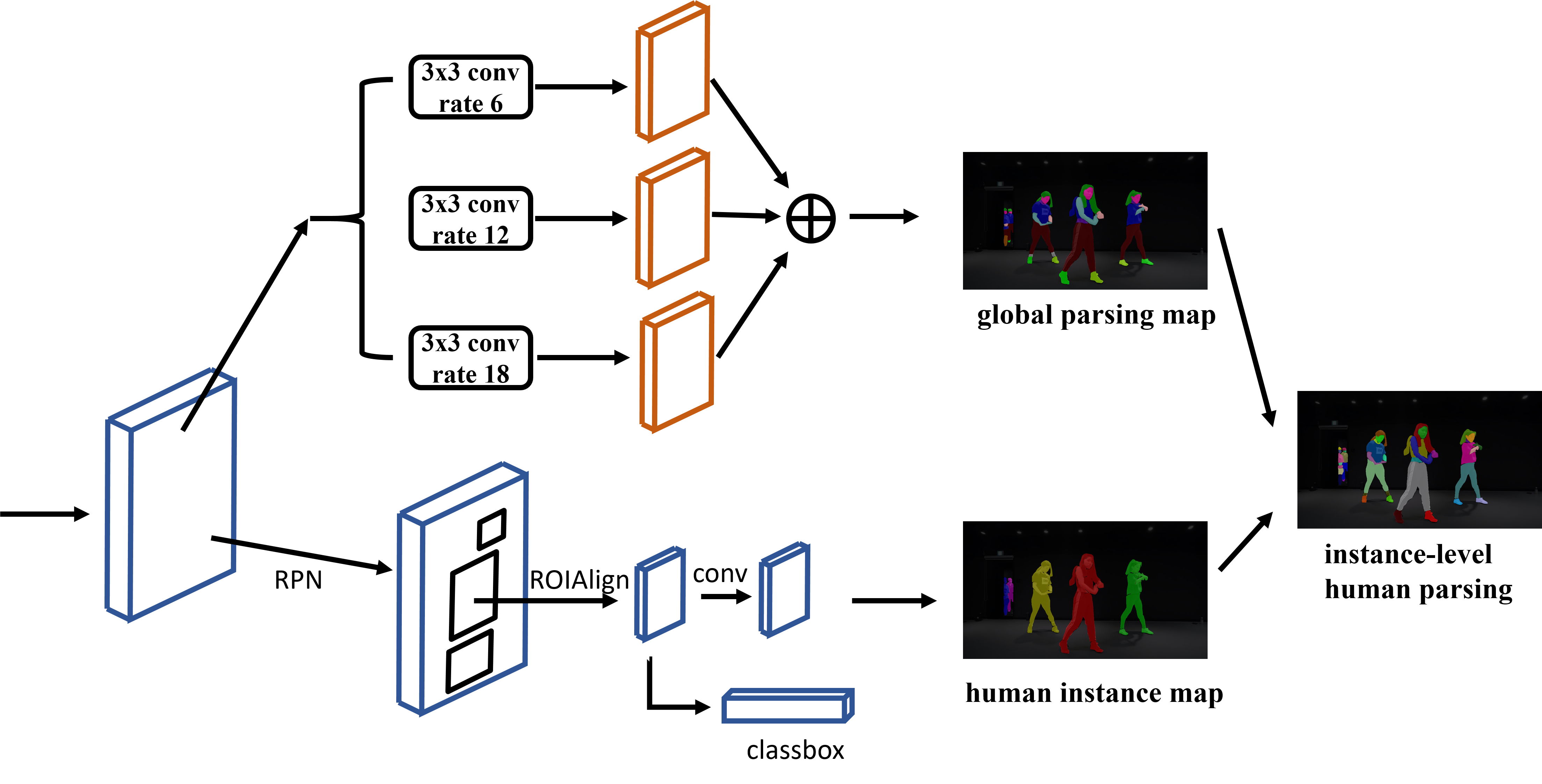}
\vspace{-4mm}
  \caption{Our Parsing-RCNN module for instance-level human parsing. Feature maps extracted by backbone network are simultaneously passed through instance-level human segmentation branch and global human parsing branch, whose results are then integrated to get final instance-level human parsing results by taking the union of all parts assigned to a particular human instance. }
\vspace{-4mm}
  \label{fig:3}
\end{figure}

\subsection{Parsing R-CNN}
\label{sess:parsingrcnn}
Following a simple and effective framework of image instance segmentation, Mask-RCNN~\cite{He_2017_ICCV}, we extend it to a Parsing-RCNN by adding a global human parsing branch to predict semantic fine-grained part segmentations in parallel with the original Mask-RCNN branch. 

As shown in Fig \ref{fig:3}, given a feature map $F$ generated by fully convolutions network\cite{long2014fully, chen2016deeplab, chen2017rethinking, Chen2018Encoder}, the whole process is as follows. On the one hand, the instance-level human segmentation branch integrates a Region Proposal Network (RPN) applied on $F$ to propose candidate object bounding boxes and ROIAlign to extract Region Of Interest (ROI) features, which performs classification, bounding-box regression and binary mask estimation. On the other hand, in the global human parsing branch, we apply multi-rate atrous convolution on $F$ to predict semantic fine-grained part segmentation, like DeepLab\cite{chen2016deeplab}. Taking these two results into consideration (human instance segmentation, semantic fine-grained part segmentation), we can easily obtain the instance-level human parsing results.

Formally, during training, we define a multi-task loss on both whole image and each ROI as 
\begin{equation}
\mathcal{L} = \mathcal{L}_{parsing} + \mathcal{L}_{cls} + \mathcal{L}_{box} + \mathcal{L}_{mask}
\label{equ:multitaskloss}
\end{equation}
$\mathcal{L}_{parsing}$ is the image global parsing loss which defined as softmax cross-entropy loss. Specially, $\mathcal{L}_{cls}$, $\mathcal{L}_{box}$ and $\mathcal{L}_{mask}$ are calculated on each ROI. Both global parsing branch and instance-level human segmentation branch are join trained to minimize $\mathcal{L}$ by stochastic gradient descent(SGD).

\subsection{Training and Inference}
\label{sess:trainandinfer}

\textbf{Training}. Our ATEN is fully differentiable and can be trained end-to-end. A standard image domain method can be transferred to video tasks by selecting a proper task-specific sub-network. 
During training phase, in each mini-batch, video frames $\{I_{k-p}, ..., I_k, I_t\}$, $-{\lfloor}l/2{\rfloor} \leq t-k < l-{\lfloor}l/2{\rfloor}$ are randomly sampled and fed to network. In the forward pass, $\mathcal{N}_{feat}$ is applied on frames except for $I_t$. After getting the encoded feature $\overline{F}_k$ as illustrated in Session \ref{sess:ate}, feature $\overline{F}_k$ is propagated to $\overline{F}_t$. Otherwise, the feature maps are identical and passed through $\mathcal{N}_{parse}$ directly. Finally, $\mathcal{N}_{parse}$ is applied on $\overline{F}_t$ or $\overline{F}_k$. Since all the components are differentiable, the multi-task loss as illustrated in Equation.\ref{equ:multitaskloss} can propagate back to all sub-networks to  optimize task performance.

\textbf{Inference}. Algorithm \ref{alg:one} summarizes the inference algorithm. Given a video frames sequence ${I}$, a segment length $l$ and an encoding rang $p$, the proposed method sequentially process each segment. Only one frame is selected as key frame in each segment. A fully convolutional network is applied on key frame $I_k$ to extract feature $F_k$. Then it searches the $p$ former key frames and feeds them into Adaptive Temporal Encoding module with the current key frame. When there are not enough former key frames, the $p$ latter key frames are selected instead. Subsequently, these key frames are warped to the current key frame via flow-guided propagation module and consecutively fed to convGRU for temporal coherence feature encoding. With the encoded feature $\overline{F}_k$, other non-key frames $\overline{F}_{t}$ feature in this segment can be obtained by flow-guided feature propagation module. Finally, Parsing-RCNN module is applied on $\overline{F}_k or \overline{F}_t$ to get instance-level parsing results.

As for runtime complexity, the ratio of our method versus the single-frame baseline is as
\begin{equation}
\small
r=\frac{O(GRU) + (l + p) \times (O(\mathcal{W}) + O({\mathcal{F}})) + l \times O(\mathcal{N}_{parse}) + O(\mathcal{N}_{feat})}{l \times (O(\mathcal{N}_{feat}) + O(\mathcal{N}_{parse}))}
\label{complex:rate}
\end{equation}
where $O(\centerdot)$ measures the function complexity. In each segment of length l, compared with frame-level baseline taking $l$ times $\mathcal{N}_{feat}$ and $\mathcal{N}_{parse}$, our method takes only one times costly $\mathcal{N}_{feat}$. As both $\mathcal{N}_{feat}$ and ${\mathcal{F}}$ have considerable complexity, we have \\$O(GRU),O(\mathcal{W}),O(\mathcal{N}_{parse}) {\ll} O({\mathcal{F}})<O(\mathcal{N}_{feat})$

Thus, the ratio in Equation.~\ref{complex:rate} is approximated as
\begin{equation}
r=\frac{(l + p) \times O({\mathcal{F}})}{l \times \mathcal{N}_{feat}} + \frac{1}{l} < 1
\end{equation}

Actually, the encoding range $p$ is smaller(\eg, 1, 2) and the backbone fully convolutional network has higher time complexity than FlowNet. Our approach with high accuracy achieves a faster speed than per-frame baseline.

\begin{algorithm}[t]
\SetAlgoNoLine
\caption{Inference algorithm of adaptive temporal encoding network}
\KwIn{video frames sequence $\{I\}$, key frame duration length $l$, encoding rang $p$}
\For{$k$ in $[1,N]$}{
  $F_{k}$ = $\mathcal{N}_{feat}(I_k)$            \Comment{extract key frame features}
}
\For{$k$ in $[1,N]$}{

  \For{$i$ in $[1, p]$}{
    $F_{k-i{\to}k} = \mathcal{W}(F_{k-i}, {\mathcal{F}}(I_k, I_{k-i})){\odot}S$
  }
    $\overline{F}_{k} = convGRU(F_{k-p{\to}k}, ..., F_{k-1{\to}k}, F_{k})$ \\
    $ $ \Comment{temporal encoding}\\
  \For{$j$ in $[-{\lfloor}l/2{\rfloor},l-{\lfloor}l/2{\rfloor})$}{ 
    \eIf{$j=0$} {
      $r_{k}$ = $\mathcal{N}_{parse}(\overline{F}_{k})$          \Comment{task-specific sub-netwok}
    } {
      $\overline{F}_{k+j}$ = $\mathcal{W}(\overline{F}_{k}, {\mathcal{F}}(I_{k+j}, I_{k})){\odot}S$\Comment{flow warping}\\
      $r_{k+j}$ = $\mathcal{N}_{parse}(\overline{F}_{k+j})$\Comment{task-specific sub-netwok}
    }
  }
}
\KwOut{instance-level human parsing results $\{r\}$}
\label{alg:one}
\end{algorithm}

\subsection{Network Architectures}
\textbf{Flow network}. We use FlowNet\cite{Dosovitskiy2015FlowNet} (simple version) as flow estimation function and join train it with other components. It is pre-trained on the Flying Chairs dataset\citep{Dosovitskiy2015FlowNet}. An extra $1\times1$ convolution layer is added to the top feature maps of FlowNet to predict scale filed, which helps to refine flow warping results. The channel of scale filed maps is equal to the dimensions of fully convolutional network output feature. In our approach, as the fully convolutional network has an output stride of 4 (see below) which is compatible to the output stride of FlowNet, we apply the image with original resolution on FlowNet and has an output stride of 4. Then flow filed and scale filed are directly used for flow warping without downsampling.

\textbf{Fully convolutional network}. We adopt the state-of-art fully convolutional network, DeeplabV3+ model\cite{Chen2018Encoder}, as the feature extractor. DeeplabV3+\cite{Chen2018Encoder} is a fully convolutional network with Atrous Spatial Pyramid Pooling module and encoder-decoder structure, which further captures multi-scale context. We use ResNet-base version of DeeplabV3+ model, which has an output stride of 4 and feature map dimension of 256.  

\textbf{Temporal encoding network}. As shown in Fig \ref{fig:2}, temporal encoding network is implemented with conGRU which is a recurrent network. As all operations in conGRU are convolution or element-wise operation, which is insensitive to input scale, our conGRU layer is able to accept the feature with arbitrary scale. Specially, several temporal correlative feature maps with the dimension of 256 are fed into conGRU layer. All convolution operations in convGRU have $3\times3$ kernel size. The output feature shape is the same as input feature, which is further passed through a  task-specific network to obtain final results.

\textbf{Parsing-RCNN}. As shown in Fig.~\ref{fig:3}, We use Mask-RCNN\cite{He_2017_ICCV} as our instance-level human segmentation branch. On the top of the 256-d feature maps from backbone fully convolutional network, the RPN sub-network and the ROI process sub-network are applied. 15 anchors (5 scales and 3 aspect ratios) are utilized in RPN, and 128 proposals are produced for each image. Besides, we extend global human parsing branch on the top of the 256-d feature maps to predict semantic fine-grained part segmentation, which contains 3 atrous convolution layers in parallel with different rate(6, 12, 18).

\begin{table}
\centering
\footnotesize
\caption{Performance comparison of part segmentation (IoU), human instance segmentation ($AP^{r}$), instance-level human parsing ($AP^{r}_{vol}$) and runtime(fps) on the VIP.}
\vspace{-2mm}
\tabcolsep 0.03in 
\begin{tabular}{c|c|cccc|cccc|c}
\toprule[0.7pt]
\multirow{2}{*}{Method}        & \multirow{2}{*}{Mean IoU} 
                               & \multicolumn{3}{c}{IoU threshold} &  \multirow{2}{*}{$AP^r$}
                               & \multicolumn{3}{c}{IoU threshold} &  \multirow{2}{*}{$AP^r_\text{vol}$}
                               & \multirow{2}{*}{fps}   \\ 
                                &   & 0.5 & 0.6 & 0.7 &  & 0.5 & 0.6 & 0.7 &  &  \\ \hline      
frame                     & 36.4 & 89.2 & 85.0 & 72.9 & 51.3 & 21.6 & 16.0 & 10.4 & 21.6 & - \\ \hline
DFF\cite{Zhu2017Deep}     & 35.3 & 89.9 & 86.4 & 74.1 & 53.2 & 20.3 & 15.0 & 9.8 & 20.3 & \textbf{8.2} \\ \hline
FGFA\cite{Zhu2017Flow}    & 37.5 & 90.6 & 88.5 & 80.2 & 57.9 & 24.0 & 17.8 & 12.2 & 23.0 & 0.8 \\ \hline
ATEN(LSTM)                & 37.9 & 90.7 & 86.3 & 81.0 & 59.8 & 24.9 & 18.7 & 12.4 & 23.9 & 3.5 \\ \hline
ATEN($p$ = 1)             & 37.3 & \textbf{90.9} & \textbf{88.8} & 81.4 & 59.6 & 24.6 & 18.4 & 12.2 & 23.6 & 4.1 \\ \hline
ATEN                      & \textbf{37.9} & 90.8 & 86.7 & \textbf{81.6} & \textbf{59.9} & \textbf{25.1} & \textbf{18.9} & \textbf{12.8} & \textbf{24.1} & 3.8 \\
\toprule[0.7pt]
\end{tabular}
\vspace{-2mm}
\label{table:vip}
\end{table}

\begin{table*}
\caption{Performance comparisons with the state-of-the-art methods on the DAVIS benchmark with intersection over union (J) and F-measure (F).}
\vspace{-2mm}
\small
\tabcolsep 0.03in 
\begin{tabular}{c|c|cccccc|cccc}
  \hline
  \multicolumn{2}{c|}{Measure} & CUT\cite{Keuper2015Motion} & FST\cite{Papazoglou2014Fast} & MP-Net-F\cite{TokmakovCVPR2017} & VM w/o CRF\cite{TokmakovICCV2017} & DFF\cite{Zhu2017Deep} & FGFA\cite{Zhu2017Flow} & ATEN-F & ATEN($l$=5) & ATEN($l$=4) & ATEN($l$=3)\\
  \hline
  \multirow{3}{*}{J} & Mean  & 55.2 & 55.8 & 70.0 & 70.1 & 67.1 & 71.2 & 69.5 & 70.2 & 70.9 & \textbf{71.3}\\
  \cline{2-12}
  & Recall & 57.5 & 64.9 & \textbf{85.0} & - & 79.0 & 82.5 & 80.5 & 84.1 & 83.9 & 84.3\\
  \cline{2-12}
  & Decay & 2.3 & 0 & 1.4 & - & 0 & 1.2 & 0.9 & 2.4 & 2.7 & 2.1\\
  \hline
  \multirow{3}{*}{F} & Mean & 55.2 & 51.1 & 65.9 & - & 65.4 & 71.0 & 69.5 & 69.0 & 71.0 & \textbf{71.5}\\
  \cline{2-12}
  & Recall & 61.0 & 51.6 & 79.2 & - & 74.9 & \textbf{82.0} & 78.4 & 79.2 & 81.3 & 81.8\\
  \cline{2-12}
  & Decay & 3.4 & 2.9 & 2.5 & - & -0.024 & 1.0 & 1.0 & 2.2 & 1.9 & 1.4\\
  \hline
\end{tabular}
\vspace{-4mm}
\label{table:davis}
\end{table*}

\section{Experiments}
\subsection{Experimental Setup}
\textbf{Dataset}. We perform detailed experimental analysis on two large-scale datasets. We first use our collected video human parsing dataset (VIP) to evaluate the performance of our approach on video instance-level human parsing. We also conduct extensive experiments on the most popular video segmentation benchmark (DAVIS)~\cite{perazzi2016benchmark} to further illustrate the superiority of our framework. 

VIP dataset contains 404 video and over 20k pixel-wise annotated images. We selected 354 videos to train our model and conduct inference on another 50 videos. We adopt standard intersection over union (IoU) criterion for evaluation of global human parsing. Following \cite{He_2017_ICCV}, we used the mean value of several mean Average Precision(mAp) with IOU thresholds from 0.5 to 0.95 for evaluation of human instance segmentation, referred as $AP^{r}$. Similarly, following \cite{li2017holistic}, the mean of the mAp scores for overlap thresholds varying from 0.1 to 0.9 in increments of 0.1, noted as $AP^{r}_{vol}$ is used to evaluate instance-level human parsing.

DAVIS~\cite{perazzi2016benchmark} contains 50 full HD videos with dense pixel-wise annotation in all frames. Only the salient objects in video clip are annotated. Following the 30/20 training/validation split provided by the dataset, we train the models on the 30 sequences and test them on 20 validation videos. We also follow the standard protocol~\cite{perazzi2016benchmark} for evaluation and report intersection over union, F-measure for contour accuracy.

\textbf{Implement Details}\footnote{The code is available at https://github.com/HCPLab-SYSU/ATEN}.
We use the basic structure and network settings provided by Deeplabv3+\cite{Chen2018Encoder}. As illustrated in Session \ref{sess:methodoverview}, key frame duration length $l$ is a fixed interval that used to set key frame and encoding range $p$ denotes how many former key frames are used for temporal encoding, which are set to 3 and 2, respectively. For VIP dataset, we use ResNet-50 as the backbone network. For both training and testing, input frames are resized to have the larger side of 512 pixels for both the feature network and the flow network. RPN uses 15 anchors and generates 128 proposals for each image. The whole framework is trained end-to-end with SGD, where 40 epochs are performed on 4 NVIDIA GeForce GTX 1080 GPUs with 8GB memory. The learning rate is $10^{-3}$ for the first 20 epochs, then reduced to $10^-4$ for the remaining 20 epochs. For both training and inference, no any bells-and-whistles like multi-scale training and box-level post-processing are used.

For DAVIS dataset~\cite{perazzi2016benchmark}, we use ResNet-101 as the backbone network. To adapt our framework to video object segmentation task, we degrade Parsing-RCNN to two class(foreground and background) semantic segmentation by removing instance-level human segmentation branch. The $384 \times 640$ inputs are randomly cropped from the frames during training. Video frames are resized to $512 \times 896$ during inference without multi-scale fusion. The whole framework is trained end-to-end with SGD, where 50 epochs are performed on 4 NVIDIA GeForce GTX 1080 GPUs with 8GB memory. The learning rate is $10^{-3}$ for the first 20 epochs and reduced to $10^-4$ for the remaining 30 epochs.

\subsection{VIP Dataset}

As we make the first attempts to solve video instance-level human parsing task, there are no existing prior methods as comparisons to our approach on the VIP dataset. We thus use the per-frame Parsing-RCNN as the basic frame-level baseline (see Table~\ref{table:vip} ``frame''). Additionally, we reimplemented DFF\cite{Zhu2017Deep} with Parsing-RCNN task-specific sub-network, which is a fast and basic feature-level method for video recognition. All the settings are the same as its paper (Key frame duration length is 10. FlowNet is used for embedding flow estimation function and jointly trained end-to-end). As shown in Table~\ref{table:vip}, DFF has a faster speed (approximately 8 fps), but fails to have a correspondingly high accuracy (even lower than frame baseline). Our approach has the best performance in both speed (approximately 4 fps) and accuracy, which performs adaptive temporal encoding over key frames and flow-guided feature propagation for consecutive frames among key frames.

\subsection{DAVIS Dataset}
\label{subsess:davis}
We compare our method with against 4 recent state-of-art methods on the unsupervised setting of the DAVIS benchmark where no manually-annotated first frame is available for the test video. For fair comparison, we only compare with the version of very recent work~\cite{TokmakovICCV2017} without using CRF post-process as our model does. Besides, we also reimplement two popular feature-level methods, DFF\cite{Zhu2017Deep} and FGFA\cite{Zhu2017Flow} on DAVIS dataset. For faster inference, FGFA only aggregates the nearest 10 frames, which is a time-consuming method aggregating features extracted from every frame.

The results are summarized in Table~\ref{table:davis}. Our ATEN achieves $1.2\%$ higher accuracy in terms of mean IoU and large improvements in speed for avoiding dense frame process, compared with the latest work~\cite{TokmakovICCV2017}. Our approach also achieves almost $2\%$ higher accuracy in terms of mean IoU and $3\times$ faster speed than the per-frame baseline (Table~\ref{table:davis} ``ATEN-F''). Note that DFF leads to $10\times$ speedup with over $2\%$ accuracy loss. Compared with Dense Feature Aggregation(FGFA)\cite{Zhu2017Flow} that performs dense feature aggregation on every frame, our ATEN extracts deep features only on sparse key frames and thus leads to a higher accuracy and much faster speed. The superior performance demonstrates that our ATEN achieves a good balance between frame-level accuracy and time efficiency by the adaptive temporal encoding and flow-guided feature Propagation.

\begin{figure*}[t]
  \centering
  \includegraphics[width=1.0\linewidth]{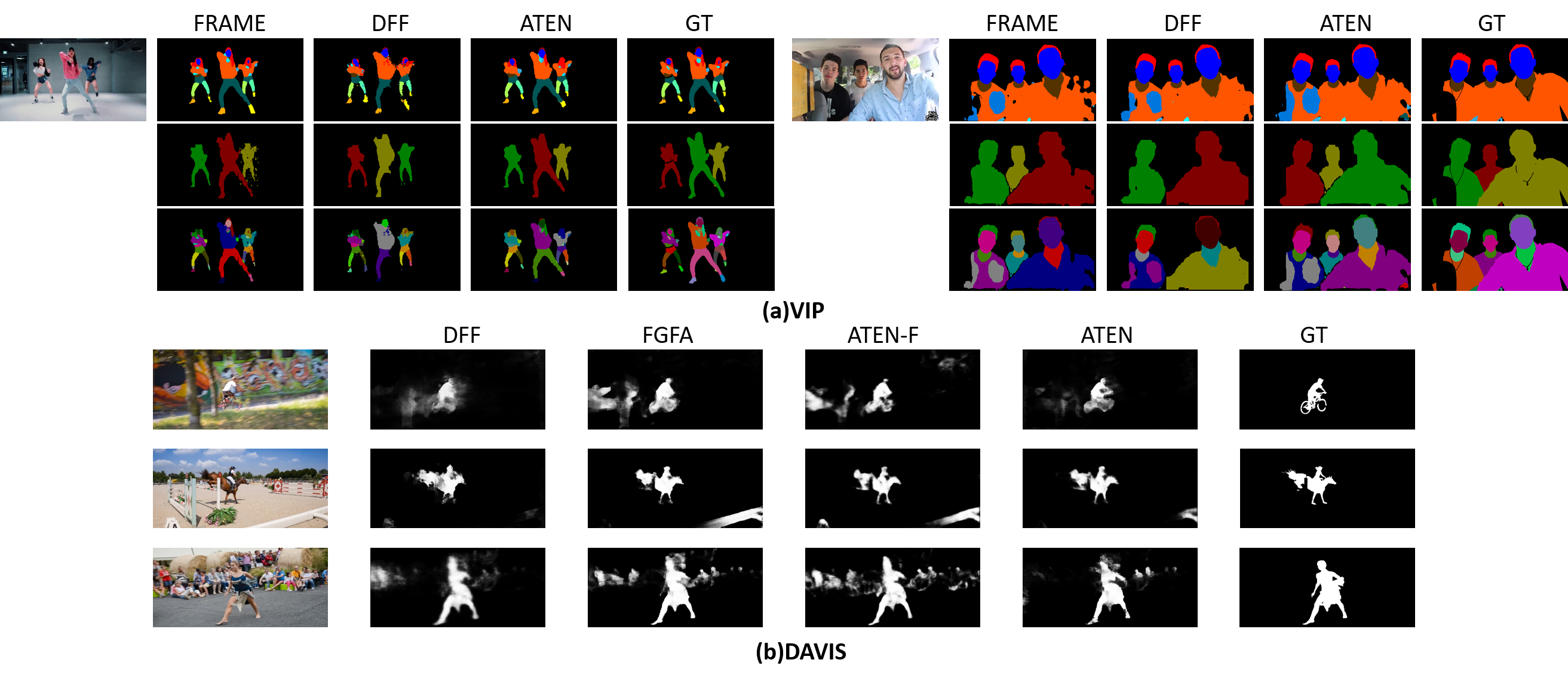}

\vspace{-4mm}
\caption{Qualitative comparisons. (a) The predicted results and ground-truth of global human parsing, instance-level human segmentation and instance-level human parsing on the VIP dataset are presented vertically. (b) Visualized comparison of video object segmentation results on the DAVIS validation set.}
\vspace{-2mm}

\label{fig:vis_results}
\end{figure*}

\subsection{Ablation Studies}
\label{subsess:ablation}

We further evaluate the effect of the main components of our ATEN as shown in Table \ref{table:ablation}. For clear comparison, in all experiments of ATEN in Table \ref{table:ablation}, encoding range $p$ is 2 and key frame duration length $l$ is 4.

\textbf{The affect of key frame duration length $l$.} We first explore the influence of the key frame duration length $l$ that actually controls the speed-accuracy trade-off, as shown in Table~\ref{table:davis}. The mean IoU decreases along with the key frame duration length increasing because large $l$ leads to highly diverse feature response between two consecutive key frames. However, larger key frame duration length leads to faster speed. When $l$ is set to 5, compared with per-frame baseline, our approach still achieves $0.7\%$ higher accuracy of mean IoU but $5\times$ faster speed.

\textbf{Flow-warping alignment}. In our ATEN as shown in Fig.~\ref{fig:2}, former key frames are warped to the current key frame via flow-guided warping and then fed into convGRU according to temporal order. The comparisons between ATEN (w/o flow-warping align) and ATEN shows that flow-warping align helps conGRU encoder learn to exploit feature of previous key frames to improve the temporal coherence of the reference key frame better. 

\textbf{Temporal encoding module}. Besides, we demonstrate that the temporal encoding module with conGRU plays an important role in our ATEN (as shown in Table~\ref{table:vip} and Table~\ref{table:ablation}). ATEN(w/o convGRU) refers to the model without convGRU encoder, which is replaced by the average of key frame features. ATEN(convLSTM) refers to the model with convLSTM. Compared to our complete ATEN, ATEN(w/o convGRU) leads to $1.7\%$ accuracy loss, which indicates that convGRU can learn to exploit temporal changes over a series of key frames and enhance the key frames feature. We also explore the performance of other recurrent neural networks by replacing convGRU with convLSTM and find that they obtain comparable results on both intersection over union(J-measures) and F-measure. 

\begin{table}
\caption{Performance comparison of different components of our ATEN on the DAVIS validation set.}
\vspace{-2mm}
\begin{tabular}{c|c}
  \hline
  Method & mIoU\\
  \hline
  ATEN(w/o flow-warping align) & 69.9\\
  \hline
  ATEN(w/o convGRU) & 69.2\\
  \hline
  ATEN(convLSTM) & 70.7\\
  \hline
  ATEN & \textbf{70.9}\\
  \hline
\end{tabular}
\vspace{-6mm}
\label{table:ablation}
\end{table}

\subsection{Qualitative Results}
To gain insights on how precise the results predicted by our ATEN, we compare the qualitative results with the baseline models on the VIP dataset and DAVIS dataset, as visualized on Fig.~\ref{fig:vis_results}. On the VIP dataset, we present the results of global human parsing, instance-level human segmentation and instance-level human parsing. By adaptively encoding temporal information, our approach demonstrates much higher precision, whereas the baseline methods incorrectly assigns some difficult body parts (\eg, the left leg of the girl in the first image) or miss some instances (\eg, the middle person in the second image). On the DAVIS dataset, we compare our model against three baseline methods, including DFF\cite{Zhu2017Deep}, FGFA\cite{Zhu2017Flow} and ATEN-F. As can be seen, the baseline models cannot handle videos with dramatic changes in appearance, like object or camera motion. It is particularly noteworthy that learning to improve the original feature with temporal coherence, our proposed approach can eventually optimize and produce better results. In general, our method generates much more accurate and consistent salient maps even in various challenging cases.

\section{Conclusions}
In this work, we investigate video instance-level human parsing that is a more pioneering and realistic task in analyzing human in the wild. We propose a novel Adaptive Temporal Encoding Network (ATEN) that alternatively performs temporal encoding network among key frames and flow-guided feature propagation for consistent frames, which achieves both high frame-level accuracy and time efficiency. To fill the blank of video human parsing data resources, we further introduce a large-scale video instance-level human parsing dataset (VIP), including 404 sequences and over 20k frames with instance-level and pixel-wise annotations. Experimental results on DAVIS~\cite{perazzi2016benchmark} and our VIP dataset demonstrate the superiority of our proposed approach, which achieves state-of-the-art performance on both video instance-level human parsing and video segmentation tasks.

\bibliographystyle{ACM-Reference-Format}
\bibliography{sigproc} 

\end{document}